\documentclass[preprint,12pt]{elsarticle}

\usepackage{amssymb}

\journal{Knowledge-Based Systems}

\usepackage[ruled]{algorithm2e}

\usepackage{bm}
\usepackage{algcompatible}
\usepackage{amssymb}
\usepackage{graphicx}
\usepackage{caption}
\usepackage[justification=centering]{subcaption}
\usepackage{array}
\usepackage{multirow}
\usepackage{color}
\usepackage{dashrule}

\usepackage{url}
\urldef{\mailsa}\path|{alfred.hofmann, ursula.barth, ingrid.haas, frank.holzwarth,|
\urldef{\mailsb}\path|anna.kramer, leonie.kunz, christine.reiss, nicole.sator,|
\urldef{\mailsc}\path|erika.siebert-cole, peter.strasser, lncs}@springer.com|

\newtheorem{definition}{Definition}

\let\quoteOLD\quote
\def\quote{\quoteOLD\small} 

\begin{document}

\begin{frontmatter}



\title{Mining Combined Causes in Large Data Sets}

\author[label1]{Saisai Ma\corref{cor1}}
\ead{saisai.ma@mymail.unisa.edu.au}
\author[label1]{Jiuyong Li}
\author[label1]{Lin Liu}
\author{Thuc Duy Le\fnref{label1}}
\address[label1]{School of Information Technology and Mathematical Sciences, University of South Australia, Mawson Lakes, SA 5095, Australia}
\cortext[cor1]{Corresponding author. Tel.: +61 451981205.}


\begin{abstract}
    In recent years, many methods have been developed for detecting causal relationships in observational data. Some of them have the potential to tackle large data sets. However, these methods fail to discover a combined cause, i.e. a multi-factor cause consisting of two or more component variables which individually are not causes. A straightforward approach to uncovering a combined cause is to include both individual and combined variables in the causal discovery using existing methods, but this scheme is computationally infeasible due to the huge number of combined variables. In this paper, we propose a novel approach to address this practical causal discovery problem, i.e. mining combined causes in large data sets. The experiments with both synthetic and real world data sets show that the proposed method can obtain high-quality causal discoveries with a high computational efficiency.
\end{abstract}


\end{frontmatter}


\section{Introduction} \label{sectionIntro}
    Causal relationships can reveal the causes of a phenomenon and predict the potential consequences of an action or an event \cite{Spirtes2010}. Therefore, they are more useful and reliable than statistical associations \cite{freedman1997association,freedman1999association,shaughnessy1985research}.

    In recent decades, causal inference has attracted great attentions in computer science. Causal Bayesian networks (CBNs) \cite{pearl2000causality,neapolitan2004learning,heckerman1995bayesian} have emerged as a main framework for representing causal relationships and uncovering them in observational data. Due to the {incapability} of CBNs in coping with high dimensional data, some efficient methods were proposed for local causal discovery around a target variable \cite{aliferis2010local,tsamardinos2003time,pellet2008using}.

    One limitation of current causal discovery methods is that they only find a cause consisting of a single variable. However, single causal factors are often insufficient for reasoning about the causes of particular effects \cite{novick2004assessing}. For example, a burning cigarette stub and inflammable material nearby can start a fire, but neither of them alone may cause a fire. With gene regulation, it was found that the expression level of a gene {might} be co-regulated by a group of other genes, which could lead to a disease \cite{wagner2007road,d2000genetic}. Furthermore, a main objective of data mining is to find previously unobserved patterns and relationships in data. Causal relationships between single variables are easier to be identified by domain experts, but {combined causes are much more difficult to be detected} \cite{mackie1965causes}. Hence data mining methods for discovering combined causes are in demand. In this paper, we address the problem of finding combined causes in large data sets.

    \begin{figure}[t]
		\centering
		\includegraphics[scale=0.7]{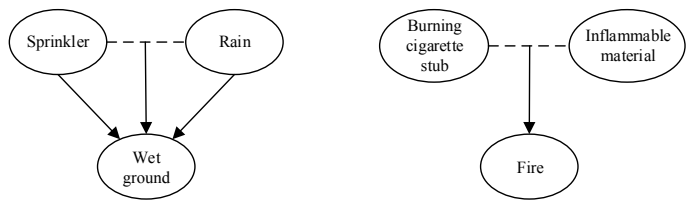}
        \caption{Multiple individual causes vs. the combined cause, where solid arrows denote causal relationships and the dashed lines represent the interaction between the two variables.}
        \label{fig:Examples}
        \vspace{-1\baselineskip}
	\end{figure}

    The combined causes considered in this paper are different from the generally discussed multiple causes. For example, in Figure \ref{fig:Examples}, sprinkler causes wet ground, and so does rain. Sprinkler and rain together cause wetter ground. However, in this work, we concern the situation when multiple variables each alone is not sufficient to cause an effect, but their  combination is. As shown in Figure \ref{fig:Examples}, there is no causal link from burning cigarette stub or inflammable material to a fire, but the combination of these two factors leads to a fire.

    The combined causes studied in this paper cannot be discovered with CBN learning, as in a CBN an edge is drawn from $A$ to $C$ only when $A$ is a cause of $C$. If $A$ and $B$ each alone is not a cause of $C$, no edge is drawn from $A$ or $B$ to $C$, and thus impossible to examine the combined causal effect of $A$ and $B$ on $C$. This limitation of CBNs was discussed in \cite{spirtes2000causation} (page 48) as follows:
    \begin{quote}
        ``Suppose drugs A and B both reduce symptoms C, but the effect of A without B is quite trivial, while the effect of B alone is not. The directed graph representations we have considered in this chapter offer no means to represent this interaction and to distinguish it from other circumstances in which A and B alone each have an effect on C.''
	\end{quote}

    To identify combined causes in data, one critical challenge is the computational complexity with large data sets, as the number of combined variables is exponential to the number of individual variables.

    In this paper, we propose a multi-level approach to discovering the combined causes of a target variable. Our method is designed based on an efficient local causal discovery method, HITON-PC \cite{aliferis2010local}, which was developed on the same theoretical ground as the well-known PC algorithm \cite{spirtes2000causation} for CBN learning.
	
    In the rest of the paper, {the related work and the contributions of this paper are described in Section \ref{sectionRelated}.} Section \ref{sectionBackg} introduces the background, including the notation and the HITON-PC algorithm. Section \ref{sectionUncover} presents the proposed method. The experiments and results are described in Section \ref{sectionExper}. Finally, Section \ref{sectionConclu} concludes the paper.

{\section{Related Work and Contributions} \label{sectionRelated}
    As discussed in the previous section, causal Bayesian networks (CBNs), as a main stream causal discovery {approach}, have been studied extensively. Many algorithms for CBN learning and inference \cite{pearl2000causality,spirtes2000causation,hill2011bayesian,mani2010bayesian} have been developed. Researchers have also tried to incorporate other models and prior knowledge into the CBN framework. The domain experts are interested in taking the prior knowledge and observational data to produce Bayesian networks \cite{heckerman1995learning}. Messaoud et al. \cite{messaoud2015semcado} proposed a framework to learn CBNs, by incorporating semantic background knowledge provided by the domain ontology. In order to address the uncertainties resulting from incomplete and partial information, Kabir et al. \cite{kabir2015integrating} combined Bayesian belief network with data fusion model to predict the failure rate of water mains. However, these methods are designed to analyse individual causes, instead of combined causes. Moreover, it may be difficult for domain experts to elicit the CBN structure with combined causes from domain knowledge only.

    Another approach \cite{segal2002learning,Azizi2014learning} was proposed to find the relationship structures between groups of variables. Segal et al. \cite{segal2002learning} defined the module network of which each node (module) was formed by a set of variables having the same statistical behavior. They also proposed an algorithm to learn the module assignment and the module network structure. Many algorithms and applications \cite{segal2003module,Azizi2014learning} have been developed to extend the module network model. Yet et al. \cite{yet2014compatible} proposed a method for abstracting the BN structure, where they also merged nodes with similar behavior to simplify the BN structure. The modules or nodes of a module network are not the same as the combined causes defined in this paper, since the components of a combined cause do not necessarily have the similar behaviour.

    The sufficient-component {cause} model \cite{rothman1976causes,rothman2005causation} (often referred by epidemiologists) addresses the combined causes discussed in this paper. According to the model, a disease is an inevitable consequence of a minimal set of factors. However, no computational methods have been developed for finding a sufficient-component cause in observational data. Although the model and interactive causes have attracted statisticians' attentions \cite{vanderweele2008empirical,vanderweele2007identification,vanderweele2012general}, the work is at the level of theoretical discussions.
		
    Li et al. \cite{li2013mining} used the idea of retrospective cohort studies \cite{euser2009cohort} and Jin et al. \cite{jin2012discovery} applied partial association tests \cite{birch1964detection} to {discover} causal rules from association rules. While the work has initiated the concept of the combined causes, their focus was on integrating association rule mining with observational studies or traditional statistical analysis for causal discovery.

    In this paper, a novel method is proposed to discover the combined causes of the given target variable, based on the causal inference framework established for CBN learning. The contributions of this paper are summarised as follows:
	\begin{enumerate}
        \item We study the problem of mining combined causes which are different from multiple individual causes, and the problem has not been tackled by most existing methods.
        \item We develop a new method for discovering combined (and single) causes, and demonstrate its performance and efficiency by experiments with synthetic and real world data.
	\end{enumerate}
}

\section{Background} \label{sectionBackg}
    {In this section, we firstly describe the notation to be used in the paper (Section \ref{subsectionNotation}). In Section \ref{subsectionHITON}, we introduce the HITON-PC algorithm, which is the basis of our algorithms, and then discuss its time complexity.}

    \subsection{Notation} \label{subsectionNotation}
        We use upper case letters, e.g. $X$ and $Y$, to represent random variables, and multiple upper case letters, e.g. $XY$ or $X\&Y$, to denote the combined variable consisting of $X$ and $Y$.  Bold-faced upper case letters, e.g. $\bm{X}$ and $\bm{Y}$, represent a set of variables. Particularly, we denote the set of predictor variables and the target variable with $\bm{V}$ and $T$ respectively. The conditional independence between $X$ and $T$ given $\bm{S}$ is represented as $I(X, T \mid \bm{S})$.

        This paper deals with binary variables only, i.e. each variable has two possible values, 1 or 0. The value of a combined (binary) variable $XY$ is 1 if and only if each of its component (binary) variables is equal to 1 (i.e. $X=1$ and $Y=1$). A multi-valued variable can be converted to a number of binary variables, e.g. the nominal variable {Eduction} can be converted to 3 binary variables, High School, Undergraduate and Postgraduate. With binary variables, we can easily create and examine a combined cause involving different values of multiple variables. For example, given the two nominal variables, Gender and {Eduction}, after converting them to binary variables, we can combine them to have variables, such as (Male, High School) and (Female, Postgraduate).

    \subsection{HITON-PC} \label{subsectionHITON}
        {Given its high efficiency and origin in the sound CBN learning theory, HITON-PC \cite{aliferis2010local} is a commonly used method for discovering local causal structures with a fixed target variable. The semi-interleaved HITON-PC is used as the basis for our proposed method.} Under the causal assumptions \cite{spirtes2000causation}, HITON-PC uses conditional independence (CI) tests to find the causal relationships around a target variable $T$, i.e. the set of parents (P) and children (C) of $T$.
				
        Referring to Algorithm \ref{alg:HITON-PC}, HITON-PC takes a data set of the predictors $\bm{V}$ and the target $T$ to produce $\emph{TPC}(T)$, the set of parents and children of $T$. The algorithm uses two data structures, a priority queue \emph{OPEN} and a list $\emph{TPC}(T)$. Initially \emph{OPEN} contains all predictors associated with $T$ and $\emph{TPC}(T)$ is empty (see lines 1 and 2 of Algorithm \ref{alg:HITON-PC}). It then iterates between the two phases, inclusion and elimination, until \emph{OPEN} becomes empty.

        \begin{algorithm}[!t]
        \caption{The Semi-interleaved HITON-PC Algorithm \cite{aliferis2010local}} \label{alg:HITON-PC}
            \textbf{Input:} A data set $\bm{D}$ for predictor variable set $\bm{V}$ and target $T$ \\
            \textbf{Output:} $\emph{TPC}(T)$, the set of parents and children of $T$
            \begin{algorithmic}[1]
                \STATE $\emph{TPC}(T)$ $\leftarrow \emptyset$
                \STATE Let \emph{OPEN} contain all variables associated with $T$, sorted in descending order of strength of associations.
                \REPEAT
                    \STATEx \hspace{\algorithmicindent} \textbf{Phase I:} Inclusion (line 4)
                    \STATE Move the first variable $X$ from \emph{OPEN} to the end of $\emph{TPC}(T)$
                    \STATEx \hspace{\algorithmicindent} \textbf{Phase II:} Elimination (lines 5-16)
                    \STATEx \hspace{\algorithmicindent} // \emph{Forward stage}
                    \IF{\emph{OPEN} $\neq \emptyset$}
                        \STATE $X \leftarrow$ the variable last added to $\emph{TPC}(T)$
                        \IF{$\exists \bm{S} \subseteq \emph{TPC}(T) \backslash \{X\}$, s.t. $I(X, T \mid \bm{S})$}
                            \STATE Remove $X$ from $\emph{TPC}(T)$
                        \ENDIF
                    \STATEx \hspace{\algorithmicindent} // \emph{Backward stage}
                    \ELSE
                        \FOR{each $X \in \emph{TPC}(T)$}
                            \IF{$\exists \bm{S} \subseteq \emph{TPC}(T) \backslash \{X\}$, s.t. $I(X, T \mid \bm{S})$}
                                \STATE Remove $X$ from $\emph{TPC}(T)$
                            \ENDIF
                        \ENDFOR
                    \ENDIF
                \UNTIL{\emph{OPEN} $= \emptyset$}
                \STATE Output $\emph{TPC}(T)$
            \end{algorithmic}
        \end{algorithm}

        In the inclusion phase, the variable having the strongest association with $T$ is removed from \emph{OPEN} and added to $\emph{TPC}(T)$ (line 4). In the elimination phase, if \emph{OPEN} is not empty, the forward stage (lines 5-9) is executed. The variable newly added to $\emph{TPC}(T)$, $X$, is eliminated from $\emph{TPC}(T)$ if it is independent of $T$ given a subset of current $\emph{TPC}(T)$, otherwise it is kept (still tentatively) in $\emph{TPC}(T)$. If \emph{OPEN} is empty, the backward stage (lines 10-16) is activated, and each variable $X$ in current $\emph{TPC}(T)$ is tested, and if a subset of $\emph{TPC}(T)$ is found such that $X$ is independent of $T$ given the subset, $X$ is removed from $\emph{TPC}(T)$.

        HITON-PC uses several heuristics to improve efficiency. At the forward stage, CI tests are conducted only on the newly added variable, instead of performing a full variable elimination. To compensate for possible false discoveries caused by this heuristic, HITON-PC uses the backward stage to ``tighten up'' $\emph{TPC}(T)$ by testing the conditional independence of each variable with $T$ given the other variables in $\emph{TPC}(T)$. Moreover, the use of the priority queue, \emph{OPEN}, allows variables having stronger associations with $T$ to be included and evaluated first. As these variables are more likely to be the true parents or children of the target, once they are in $\emph{TPC}(T)$,  it is expected that given these variables, {other variables that should not be in $\emph{TPC}(T)$ are quickly identified and removed} so that in the forward stage $\emph{TPC}(T)$ will not be over expanded, thus reducing the number of CI tests. Additionally, in practice, HITON-PC restricts the maximum order of CI tests to a given threshold \emph{max-k}, i.e. the maximum size of the conditioning set ($\bm{S}$, see Algorithm \ref{alg:HITON-PC}) is \emph{max-k}.

        The time complexity of HITON-PC mainly depends on the number of CI tests. Each variable needs to be tested on all subsets of $\emph{TPC}(T)$. Thus the complexity regarding each variable is ${O(2^{|TPC(T)|})}$, and the total time complexity is ${O(|\bm{V}|2^{|TPC(T)|})}$. When \emph{max-k} is specified, the complexity becomes polynomial, i.e. ${O(|\bm{V}| |\emph{TPC(T)}|^{max-k})}$. Extensive experiments have shown that HITON-PC is able to cope with thousands of variables with low rate of false discoveries \cite{aliferis2010local}.

\section{Uncovering Combined Causes} \label{sectionUncover}
    {Having introduced the background knowledge, in this section, we present the proposed method for discovering combined causes. We firstly introduce the na\"{i}ve approach (Section \ref{subsectionNaive}), which is a straightforward way to detect the combined causes. Then we give the formal definition of combined causes and present the basic idea of the proposed method (Section \ref{subsectionBasic}). Finally, we describe the proposed method (including two algorithms, MH-PC-F and MH-PC-B) for discovering combined causes (Section \ref{subsectionMH-PC-F}) and discuss their possible false discoveries (Section \ref{subsectionFalse}).}

    \subsection{The Na\"{i}ve Approach} \label{subsectionNaive}
        A na\"{i}ve scheme for finding the combined causes can be as follows. Firstly, we generate a new variable set with combined variables using the original variable set. For example, for $\bm{V} = \{A, B, C, D, E, X, Y, Z\}$, the new variable set (with $2^8-1$ variables) is $\bm{V'} = \{A, ..., Z,$ $\emph{AB}, ..., \emph{YZ},$ $\emph{ABC}, ..., \emph{XYZ},$ $..., \emph{ABCDEXYZ}\}$. Then we run a local causal discovery algorithm, such as HITON-PC, to find both single and combined causes using the data set created for $\bm{V'}$.
				
        The na\"{i}ve approach, however, is not feasible because the number of combined variables is exponential to the number of individual variables. In the following, we discuss how our proposed method tackles the problem.

    \subsection{Basic Idea of the Proposed Method} \label{subsectionBasic}
        In fact, it is not necessary to consider all combined variables. Particularly, we are not interested in a combined variable, {e.g. $W=XY$}, of which a component {$X$ or $Y$} is a cause already, as it is reasonable to assume that the causal relationship between $W$ and { the} target $T$ is due to the relationship between $X$ {or $Y$} and $T$. To improve efficiency, we can exclude such combined variables when finding combined causes of $T$, and only consider the combined variables whose components are not causes of $T$.

        Furthermore, as discussed in Section \ref{sectionIntro}, a combined cause consisting of non-cause components is more difficult to be observed by domain experts and they cannot be represented or discovered using other approaches such as CBNs, hence mining such combined causes is useful in practice.

        The definition of the combined causes studied in this paper is given below.
        \begin{definition}[Combined Cause] \label{defCombinedCause}
            Let $W$ be a combination of multiple variables. $W$ is a combined cause of $T$ if $W$ is a cause of $T$ and any of its component variables, $X \in W$, is not a cause of $T$.
        \end{definition}

        Based on Definition \ref{defCombinedCause}, we can design an algorithm to find such combined causes (and single causes) in a level by level manner. We firstly, at the \emph{$(k-1)^{th}$ level} ($k \geq 2$), obtain $\emph{TPC}_{k-1}(T)$, the set of parents and children of $T$, each consisting of $k-1$ individual variables. Then at the \emph{$k^{th}$ level}, combined variables are generated based on $\emph{nTPC}(T)$, the set of non-cause variables at all { the} lower levels, i.e. $\emph{nTPC}(T) = \emph{nTPC}_1(T) \cup \ldots \cup \emph{nTPC}_{k-1}(T)$, where $\emph{nTPC}_i(T)$ ($i \in \{1,\ldots,k-1\}$) is the set of non-cause variables each containing $i$ individual variables. For example, a $k^{th}$ \emph{level} combined variable can be generated by combining an $i^{th}$ \emph{level} ($i \in \{1,\ldots,k-1\}$) non-cause variable and a $(k-i)^{th}$ \emph{level} non-cause variable.

        {For non-cause variables $X$ and $Y$, if the combination $XY$ is a combined cause, then it is reasonable to assume that $X$ positively contributes to the relationship between $Y$ and the target $T$ and vice versa. For example, the combustible dust suspended in the air (even at a high concentration) has no causal effect on a dust explosion, but an ignition source will improve their relationship significantly and thus the combination of the two factors can result in a dust explosion. This observation leads to the following definition.}

        \begin{definition}[Redundant Combined Variable]
            {For $\forall X, Y \in \textbf{V}$, the combination $XY$ is a redundant combined variable, if either $I(X, T | Y=1)$ or $I(Y, T | X=1)$, where $X$ and $Y$ are not individual causes of the target $T$.}
        \end{definition}

        By excluding redundant combined variables, we can further improve the efficiency of the causal discovery.
			
        Based on the above discussion and HITON-PC, we propose the Multi-level HITON-PC (MH-PC) { method}  for finding both single and combined causes of a given target. In the following section, we present the details of the method.

        \subsection{Multi-level HITON-PC} \label{subsectionMH-PC-F}
        Referring to Algorithm \ref{alg:MH-PC}, at the first level, MH-PC invokes HITON-PC to find the single causes of $T$, $\emph{TPC}_1(T)$ (line 1) and initiates $\emph{TPC}(T)$ as $\emph{TPC}_1(T)$ (line 2). The single non-cause variables are put in $\emph{nTPC}_1(T)$ and $\emph{nTPC}(T)$ (non-cause variables identified at all { the} lower levels) is initially empty (line 2).

        \begin{algorithm}
        \caption{\textbf{M}ulti-level \textbf{H}ITON-\textbf{PC} (\textbf{MH-PC})} \label{alg:MH-PC}
            \textbf{Input:} A data set $\bm{D}$ for predictor variable set $\bm{V}$ and target $T$; $k_{max}$, the maximum level of causal discovery \\
              \textbf{Output:} \emph{TPC}$(T)$, the set of (single and combined) parents and children of $T$
            \begin{algorithmic}[1]
                \STATE Call Algorithm \ref{alg:HITON-PC} (HITON-PC), i.e. $\emph{TPC}_1(T) = $ HITON-PC($\bm{D},\bm{V},T$)
				\STATE $\emph{TPC}(T) = \emph{TPC}_1(T)$; $\emph{nTPC}_1(T) = \bm{V} \backslash \emph{TPC}_1(T)$; $\emph{nTPC}(T) = \emptyset$
                \FOR{$k = 2$ to $k_{max}$}
										\STATE \emph{nTPC}$(T) = \emph{nTPC}(T) \cup \emph{nTPC}_{k-1}(T)$
                    \STATE Generate \emph{$k^{th}$ level} combined variable set $\bm{V'}$ based on \emph{nTPC}$(T)$
                    \STATE $\bm{V_k} =$ redundancyTest($\bm{V'}$)
					\STATE Generate corresponding data set $\bm{D_k}$ for $\bm{V_k}$
                    \STATE Let \emph{OPEN} contain all variables (in $\bm{V_k}$) associated with $T$, sorted
                    \STATEx\hspace{\algorithmicindent} in descending order of strength of associations.
                \REPEAT
                    \STATEx\hspace{\algorithmicindent}\hspace{\algorithmicindent} \textbf{Phase I:} Inclusion (line 10)
                    \STATE Move the first variable from \emph{OPEN}, add it to the end of $\emph{TPC}(T)$
                    \STATEx\hspace{\algorithmicindent}\hspace{\algorithmicindent} and $\emph{TPC}_k(T)$
                    \STATEx\hspace{\algorithmicindent}\hspace{\algorithmicindent} \textbf{Phase II:} Elimination (lines 11-22)
                    \STATEx\hspace{\algorithmicindent}\hspace{\algorithmicindent} // \emph{Forward stage}
                    \IF{\emph{OPEN} $\neq \emptyset$}
                        \STATE $X \leftarrow$ the variable last added to \emph{TPC}$(T)$
                        \IF{$\exists \bm{S} \subseteq$ $\emph{TPC}(T)$, s.t. $I(X, T \mid \bm{S})$}
                            \STATE Remove $X$ from \emph{TPC}$(T)$ and \emph{TPC}$_k(T)$
                        \ENDIF
                    \STATEx\hspace{\algorithmicindent}\hspace{\algorithmicindent} // \emph{Backward stage}
                    \ELSE
                        \FOR{each $X \in \emph{TPC}_k(T)$ }
                            \IF{$\exists \bm{S} \subseteq \emph{TPC}(T) \backslash \{X\}$, s.t. $I(X, T \mid \bm{S})$}
                                \STATE Remove $X$ from \emph{TPC}$(T)$ and \emph{TPC}$_k(T)$
                            \ENDIF
                        \ENDFOR
                    \ENDIF
                \UNTIL{\emph{OPEN} $= \emptyset$}
								\STATE $\emph{nTPC}_k(T) = \bm{V_k} \backslash \emph{TPC}_k(T)$
                \ENDFOR
                \STATE Output \emph{TPC}$(T)$
            \end{algorithmic}
        \end{algorithm}
			
        At level $k$ ($k\ge 2$), MH-PC firstly updates $\emph{nTPC}(T)$ so that it contains the non-causes from levels 1 to $k-1$ (line 4). Next the algorithm generates combined variables containing $k$ individual variables by combining the variables in $\emph{nTPC}(T)$ (line 5). Redundant combined variables are then removed (line 6) and the new data set $D_k$ for level $k$ combined variables ($\bm{V_k}$) is created too (line 7). From lines 8 to 23, we identify level $k$ combined causes from $\bm{V_k}$. Initially \emph{OPEN} contains all the combined variables in $\bm{V_k}$ which are associated with $T$ and the variables are sorted in descending order of the strength of associations (line 8). Similar to HITON-PC, the inclusion and elimination phases are carried out iteratively till \emph{OPEN} is empty. At the end of the iteration (line 23), $\emph{TPC}_k(T)$ includes the discovered combined causes consisting of $k$ variables, and $\emph{TPC}(T)$ includes all the causes from level 1 to level $k$. Note that in line 17, to improve the efficiency further, the backward stage only checks the level $k$ candidates in $\emph{TPC}_k(T)$, instead of all candidates in $\emph{TPC}(T)$, as all the lower level parents and children have been confirmed at previous levels.
				
        In line 24, the set of level $k$ non-causes is updated before completing the work at level $k$. Finally MH-PC outputs $\emph{TPC}(T)$ until $k=k_{max}$ (the maximum level of causal discovery).

        In the forward stage (i.e. the case when \textit{OPEN} $\neq \emptyset$), as with HITON-PC, MH-PC searches the current $\emph{TPC}(T)$ for a subset $\bm{S}$ to test whether the combined variable $X$ is independent of $T$ given $\bm{S}$ (line 13). Since combined variables in $\emph{TPC}(T)$ are combinations of individual variables, MH-PC may have conducted some redundant conditional independence tests. For example, the CI test between $X$ and $T$ given a combined variable $\emph{YZ}$ (i.e. $I(X,T \mid {\emph{YZ}})$) may be unnecessary if the test given the two individual variables $Y$ and $Z$ (i.e. $I(X,T \mid Y,Z)$) has been done.
        To address this issue, we propose a variant of MH-PC, called \textbf{MH-PC-B} (B version of MH-PC). To avoid confusion, in the rest of the paper, we call the MH-PC algorithm shown in Algorithm \ref{alg:MH-PC} {\textbf{MH-PC-F} (Full version of MH-PC). In the forward stage, when conducting the level $k$ test with MH-PC-B, we do not include the level $k$ variables in $\emph{TPC}_k(T)$ into conditioning sets, i.e. we replace line 13 in Algorithm \ref{alg:MH-PC} with the following statement:

				$$  {\textbf{if}}\  {\exists \bm{S} \subseteq \emph{TPC}(T) {\backslash \emph{TPC}_k(T)}, s.t. \, I(X, T \mid \bm{S})} $$

        As MH-PC-B conducts the tests conditioning only on {the} lower level variables, it can have higher efficiency {than MH-PC-F}, but at the same time it may produce some false positives. However, since in the backward stage (lines 17-21 of Algorithm \ref{alg:MH-PC}), we do another check of the candidate causes remained in $\emph{TPC}_k(T)$, it is expected that the false discoveries are removed. As we will see from the next section, the experiments show that MH-PC-B is more efficient than MH-PC-F, while producing the same results as MH-PC-F with the data sets used.

    \subsection{False Discoveries of Multi-level HITON-PC} \label{subsectionFalse}
        Since MH-PC-F (and MH-PC-B) follows the idea of HITON-PC, we firstly analyse the quality of HITON-PC in term of false discoveries. In HITON-PC, possible false decisions mainly come from two sources \cite{spirtes2000causation,li2009controlling}: the use of $max$-$k$, the maximum size of conditioning sets ($\bm{S}$, see {Algorithms} \ref{alg:HITON-PC} and \ref{alg:MH-PC}) used for conditional independence tests, and incorrect results of statistical tests. Using a smaller $max$-$k$ reduces the number of conditional independence tests, thus improves efficiency, but results in false positive discoveries. Fortunately, when $max$-$k = 3$ or 4, the false positive rate is not high, as shown in \cite{aliferis2010local}. When we {do not have} enough number of samples, the statistical tests may produce incorrect results.

        In the following, we will discuss the false discoveries coming from the interactions between variables. As mentioned above, the proposed method only focuses on non-redundant combined variables (Definition 2). While this strategy is used for reducing complexity, it may lead to false discoveries. However, we argue that our algorithms can still obtain high-quality causal findings. {{For} non-cause variable $X$, if another non-cause variable $Y$ cannot improve the relationship between $X$ and the target $T$, then the combination $XY$, in most cases, may {not improve} the relationship between $X$ and $T$. This type of combined variables are unlikely to be combined causes.} The experiment results in Section \ref{sectionExper} have confirmed this intuition.

\section{Experiments} \label{sectionExper}

    We implemented MH-PC-F and MH-PC-B based on the semi-interleaved HITON-PC implementation in the R package, \emph{bnlearn} \cite{scutari2009learning}. In the experiments, the maximum level of combination (i.e. $k_{max}$ in Algorithm \ref{alg:MH-PC}) is restricted to 2, i.e. a combined cause at most consists of two component variables. {We set the threshold of $p$-value to 0.01 to prune redundant combined variables and 0.05 to test causal relationships, for both synthetic and real world data sets.}

    \subsection{Data Sets} \label{subsectionData}
        {10 synthetic and 7 real world data sets were used in the experiments, and a summary of the data sets is shown in Table \ref{table:data}.} The variables in all data sets are binary, i.e. each variable has two possible values, 1 or 0. The class variable in each data set is specified as the target variable. The numbers of variables shown in the table refer to the numbers of single predictor variables. The distribution of each data set indicates the percentages of the two different values of class variables. For synthetic data sets, the ground truth (i.e. the number of true causes) is shown in the table, where the first value is the number of single causes each consisting of one predictor variable and the second value is the number of combined causes each consisting of two predictor variables.
		
        \begin{table}[!t]
        \caption{A Brief Description of Data Sets} \label{table:data}
            \centering 
            \begin{tabular}{|c | c | c | c | c|} 
                \hline 
                    Name & \#Records & \#Variables & Distributions & Ground Truth \\ [0ex] 
                \hline 
                    Syn-7 & 1000 & 6 & 39.1\% \& 60.9\% & 2, 1 \\
                    Syn-10 & 1000 & 9 & 42.2\% \& 57.8\% & 3, 2 \\
                    Syn-12 & 2000 & 11 & 72.1\% \& 27.9\% & 4, 3 \\
                    Syn-16 & 2000 & 15 & 45.2\% \& 54.8\% & 4, 3 \\
                    Syn-20 & 2000 & 19 & 55.6\% \& 44.4\% & 4, 4 \\
                \hline
                    Syn-50 & 5000 & 49 & 30.8\% \& 69.2\% & 5, 5 \\
                    Syn-60 & 5000 & 59 & 30.8\% \& 69.2\% & 5, 5 \\
                    Syn-80 & 5000 & 79 & 30.5\% \& 69.5\% & 5, 5 \\
                    Syn-100 & 5000 & 99 & 30.0\% \& 70.0\% & 5, 5 \\
                    Syn-120 & 5000 & 119 & 30.4\% \& 69.6\% & 5, 5 \\
                \hline\hline
                    CMC & 1473 & 22 & 57.3\% \& 42.7\% & - \\
                    {German} & {1000} & {60} & {30.0\% \& 70.0\%} & {-}\\
                    House-votes-84 & 435 & 16 & 61.4\% \& 38.6\% & - \\
                    Hypothyroid & 3163 & 51 & 4.8\% \& 95.2\% & {-}\\
                    {Kr-vs-kp} & {3196} & {74} & {47.8\% \& 52.2\%} & {-}\\
                    Sick & 2800 & 58 & 6.1\% \& 93.9\% & - \\
                    Census & 299285 & 495 & 6.2\% \& 93.8\% & {-} \\ [0ex] 
                \hline 
            \end{tabular}
        \end{table}

        {The first five synthetic data sets (with small number of variables) in Table \ref{table:data} were generated with two main steps: (1) generating a data set based on a BN (Bayesian network) created randomly by the TETRAD software tool (http://www.phil.cmu.edu/tetrad/), and (2) generating the final synthetic data set by ``splitting'' some causes of the target into two new variables. Specifically, we firstly created a random BN using the TETRAD software, whose structure and conditional probability tables were both generated randomly. In the obtained BN, one of the variables was designated as the target and the others as predictor variables. Records of all variables were generated based on the conditional probability tables, using the built-in Bayes Instantiated Model. Then we selected and split a parent node, e.g. $A$, of the target into two variables, e.g. $A_1$ and $A_2$, such that (1) $A_1 \wedge A_2 = A$ (i.e. $A_1$ and $A_2$ both are equal to 1 if and only if $A$ is 1), and (2) $A_1$ or $A_2$ is not an individual cause of the target. Note that, for combined causes in the synthetic data, we do not have a complete ground truth, since it may include some combined causes that we do not observe.

        For the next five larger synthetic data sets (Syn-50, ..., Syn-120), it is {unpractical} to generate them based on randomly drawn BNs, since it takes too long time to generate one. We firstly drew a simple BN where some variables were the parents of the target, some were not. Then we adopted logistic regression to generate the data based on the BN. Next, we employed the aforementioned splitting process to obtain the final data sets.}
	
        All real world data sets shown in Table \ref{table:data} are {obtained} from the UCI Machine Learning Repository \cite{asuncion2007uci}. The first six real world data sets were employed to assess the effectiveness of the proposed algorithms, while the Census data set was used for evaluating the efficiency. The CMC (Contraceptive Method Choice) data set is an extraction of the National Indonesia Contraceptive Prevalence Survey in 1987. The German data set is a data set for classifying people's credit risks based on a set of attributes. House-votes-84 contains the United States Congressional Voting Records in 1984. Hypothyroid and Sick are two medical data sets, which are from the Thyroid Disease data set of the repository (discretised using the MLC++ discretisation utility \cite{kohavi.ea:using-mlc:96}). The Kr-vs-kp data set is generated and described based on a chess game, King-Rook versus King-Pawn on A7 (usually abbreviated {as} KRKPA7). The Census data set is the Census Income (KDD) data set from the UCI Machine Learning Repository. In our experiments, all continuous attributes have been removed from the original data sets.

    \subsection{Performance Evaluation} \label{subsectionPerformance}
        { Three} sets of experiments with the synthetic data were done to assess the accuracy of MH-PC-F and MH-PC-B by examining the results against the ground truth.

            \begin{table} [!t]
            \caption{Comparison of the proposed algorithms with the na\"{i}ve methods} \label{table:Syn}
                \centering
                \begin{tabular}{| c | c | c | c | c | c | c |}
                    \hline
                        Data & Predictor & Ground & \multirow{2}{*}{Na\"{i}ve-H} & \multirow{2}{*}{Na\"{i}ve-S} & \multirow{2}{*}{MH-PC-F} & \multirow{2}{*}{MH-PC-B} \\
                        set & variables & truth & & & & \\ [0ex]
                    \hline
                        \multirow{5}{*}{Syn-7} & $V3$ & Yes & $\surd$ & $\surd$ & $\surd$ & $\surd$ \\
                        & $V4$ & Yes & $\surd$ & $\surd$ & $\surd$ & $\surd$ \\
                        & $V1\&V2$ & Yes & $\surd$ & & $\surd$ & $\surd$ \\
                        & $V3\&V5$ & No & $\surd$ & $\surd$ & & \\
                        & $V4\&V5$ & No & $\surd$ & $\surd$ & & \\ [0ex]
                    \hline
                    \hline
                        \multirow{9}{*}{Syn-10} & $V5$ & Yes & $\surd$ & $\surd$ & $\surd$ & $\surd$ \\
                        & $V6$ & Yes & $\surd$ & $\surd$ & $\surd$ & $\surd$ \\
                        & $V7$ & Yes & & & & \\
                        & $V1\&V2$ & Yes & $\surd$ & $\surd$ & $\surd$ & $\surd$ \\
                        & $V3\&V4$ & Yes & $\surd$ & $\surd$ & $\surd$ & $\surd$ \\
                        & $V1\&V3$ & No & $\surd$ & & & \\
                        & $V4\&V5$ & No & $\surd$ & & & \\
                        & $V5\&V6$ & No & $\surd$ & $\surd$ & & \\
                        & $V5\&V9$ & No & & $\surd$ & & \\ [0ex]
                    \hline
                \end{tabular}
            \end{table}

        {Firstly we compared MH-PC-F and MH-PC-B with two na\"{i}ve approaches using HITON-PC and PC-select \cite{pcalg} respectively (denoted as Na\"{i}ve-H and Na\"{i}ve-S in the following). PC-select is an effective method for discovering the parents and children of a target variable, so we employed it as a benchmark for accuracy comparison.

        Because the two na\"{i}ve methods, especially Na\"{i}ve-S, cannot handle large data sets, two small synthetic data sets (Syn-7 and Syn-10) were used in this set of experiments. Moreover, it is easier for small data sets to provide a good visualization of the detailed results.}

        The ground truth of the Syn-7 data set is that $V3$ and $V4$ are two single causes of the target and $V1\&V2$ is a combined cause {(see the Ground truth column of Table \ref{table:Syn}, where Yes means the predictor variable is a cause of the target, and No means otherwise)}. {In Table \ref{table:Syn},} MH-PC-F and MH-PC-B find exactly the ground truth in Syn-7. While Na\"{i}ve-H identifies the ground truth, {it includes a number of redundant results, for example, $V3\&V5$ and $V4\&V5$ since $V3$ and $V4$ are causes already}. Na\"{i}ve-S misses the combined cause ($V1\&V2$) and {it finds some redundant combined causes too}. Similar results can be observed with the Syn-10 data set. MH-PC-F and MH-PC-B miss the true single cause, $V7$, {and} the na\"{i}ve methods do not find it either.

        \begin{table} [!t]
            \caption{Comparison of combined causes discovered by CR-CS, CR-PA, MH-PC-F and MH-PC-B with small synthetic data sets} \label{table:Syn-small}
            \centering
            \begin{tabular}{|c|c|c|c|c|}
                \hline
                    \multicolumn{2}{|c|}{} & Syn-12 & Syn-16 & Syn-20 \\
                \hline
                    \multirow{3}{*}{CR-CS} & $P$     & 0.25  & 0.23  & 0.36 \\
                    & $R$     & 1.00  & 1.00  & 1.00 \\
                    & $F_1$    & 0.40  & 0.38  & 0.53 \\
                \hline
                    \multirow{3}{*}{CR-PA} & $P$     & 0.16  & 0.17  & 0.27 \\
                    & $R$     & 1.00  & 1.00  & 1.00 \\
                    & $F_1$    & 0.27  & 0.29  & 0.42 \\
                \hline
                    \multirow{3}{*}{MH-PC-F} & $P$     & 0.67  & 0.50  & 1.00 \\
                    & $R$     & 0.67  & 1.00  & 1.00 \\
                    & $F_1$    & 0.67  & 0.67  & 1.00 \\
                \hline
                    \multirow{3}{*}{MH-PC-B} & $P$     & 0.67  & 0.50  & 1.00 \\
                    & $R$     & 0.67  & 1.00  & 1.00 \\
                    & $F_1$    & 0.67  & 0.67  & 1.00 \\
                \hline
            \end{tabular}
        \end{table}

        \begin{table} [!ht]
            \caption{Comparison of combined causes discovered by CR-CS, CR-PA, MH-PC-F and MH-PC-B with larger synthetic data sets} \label{table:Syn-large}
            \centering
            \begin{tabular}{|c|c|c|c|c|c|c|}
                \hline
                    \multicolumn{2}{|c|}{} & Syn-50 & Syn-60 & Syn-80 & Syn-100 & Syn-120 \\
                \hline
                    \multirow{3}[0]{*}{CR-CS} & $P$     & 0.71  & 1.00  & 0.71  & 0.83  & 1.00 \\
                    & $R$     & 1.00  & 1.00  & 1.00  & 1.00  & 1.00 \\
                    & $F_1$    & 0.83  & 1.00  & 0.83  & 0.91  & 1.00 \\
                \hline
                    \multirow{3}[0]{*}{CR-PA} & $P$     & 0.71  & 1.00  & 0.83  & 0.83  & 1.00 \\
                    & $R$     & 1.00  & 1.00  & 1.00  & 1.00  & 1.00 \\
                    & $F_1$    & 0.83  & 1.00  & 0.91  & 0.91  & 1.00 \\
                \hline
                    \multirow{3}[0]{*}{MH-PC-F} & $P$     & 0.83  & 0.71  & 1.00  & 0.83  & 0.83 \\
                    & $R$     & 1.00  & 1.00  & 1.00  & 1.00  & 1.00 \\
                    & $F_1$    & 0.91  & 0.83  & 1.00  & 0.91  & 0.91 \\
                \hline
                    \multirow{3}[0]{*}{MH-PC-B} & $P$     & 0.83  & 0.71  & 1.00  & 0.83  & 0.83 \\
                    & $R$     & 1.00  & 1.00  & 1.00  & 1.00  & 1.00 \\
                    & $F_1$    & 0.91  & 0.83  & 1.00  & 0.91  & 0.91 \\
                \hline
            \end{tabular}%
        \end{table}

        {Then we compared MH-PC-F and MH-PC-B with CR-CS \cite{li2015from} and CR-PA \cite{jin2012discovery} using three synthetic data sets, Syn-12, Syn-16 and Syn-20. CR-CS and CR-PA are both designed to explore causal relationships from association rules, and they are also capable of finding both single and combined causes. The results are shown in Table \ref{table:Syn-small}, where $P$, $R$ and $F_1$ represent the Precision, Recall and $F_1$-measure respectively. In the paper, we used odds ratio greater than 1.5 as the {threshold} to indicate a significant result in both CR-CS and CR-PA. We can see that MH-PC-F and MH-PC-B both achieve higher accuracy than CR-CS and CR-PA, based on the known ground truth. Actually, CR-CS and CR-PA both perform very well in term of Recall, but they also include many false positives, since a main aim of these two methods is for explorations and they tolerate false positives and seek high recall.}

        In the {next} set of experiments, {the last five larger} synthetic data sets in Table \ref{table:data} were used. From Table \ref{table:Syn-large}, {all the four algorithms (i.e. CR-CS, CR-PA, MH-PC-F and MH-PC-B) can recover the ground truth very well from the data sets with relatively large sizes.}

        \begin{table}[!t]
            \centering
            \caption{Number of (single and combined) causes discovered by MH-PC-F and MH-PC-B in real world data sets}
            \begin{tabular}{|c|c|c|}
                \hline
                    & No. of single causes & No. of combined causes \\
                \hline
                    CMC & 2 &  0 \\
                    German & 1 & 13 \\
                    House-votes-84 & 0 & 18 \\
                    Hypothyroid & 3 & 4 \\
                    Kr-vs-kp & 7 & 12 \\
                    Sick & 3 & 8 \\
                \hline
            \end{tabular}%
            \label{table_number}%
        \end{table}%

        Based on the results of {three} sets of experiments{,} it is reasonable to conclude that MH-PC-F and MH-PC-B are {capable to find single and combined causes}. Another finding is that the causes (single and combined) identified by MH-PC-F and MH-PC-B are always the same, and this indicates two algorithms can achieve consistent results. This is also demonstrated by the results of two algorithms with all real world data sets, as described in the following.

        To investigate combined causes in the real world cases, we ran the proposed algorithms on the first six real world data sets in Table \ref{table:data} for performance evaluation, where MH-PC-F and MH-PC-B still return consistent results as shown in Table \ref{table_number}. The proposed algorithms find many combined causes, and some of the combined causes discovered are reasonable as judged by common sense, shown in Table \ref{table:ExmComCauses}. {For example, from the Sick data set it is found that a low level of TT4 (Total T4) and T3 may result in thyroid disease (Table \ref{table:ExmComCauses}, where T4 and T3 are hormones produced by thyroid), and being sick and having a low level of T3 can lead to thyroid disease too. Some interesting combined causes are also discovered in the German data set. If one person has a private real estate and does not apply for any other installment plan, then this person is very likely to have a low default risk.}

        \begin{table} [!t]
            \caption{Examples of combined causes identified from Sick and German data sets} \label{table:ExmComCauses}
            \centering
            \begin{tabular}{c | c}
                \hline
                    \multirow{2}{*}{Sick} & T3 $< 1.151$ \& TT4 $< 87.5$ \\
                    & sick = true \& T3 $< 1.151$ \\
                \hline\hline
                    \multirow{2}{*}{German} & Checking.account = no-account \& Savings.account $<$ 100DM \\
                \cline{2-2}
                    & Property = real.estate \& Other.installment.plans = none \\
                \hline
            \end{tabular}
        \end{table}

    \subsection{Efficiency and Scalability} \label{subsectionEfficiency}
        \begin{figure}[!t]
            \begin{center}
            \includegraphics[width=0.65\textwidth]{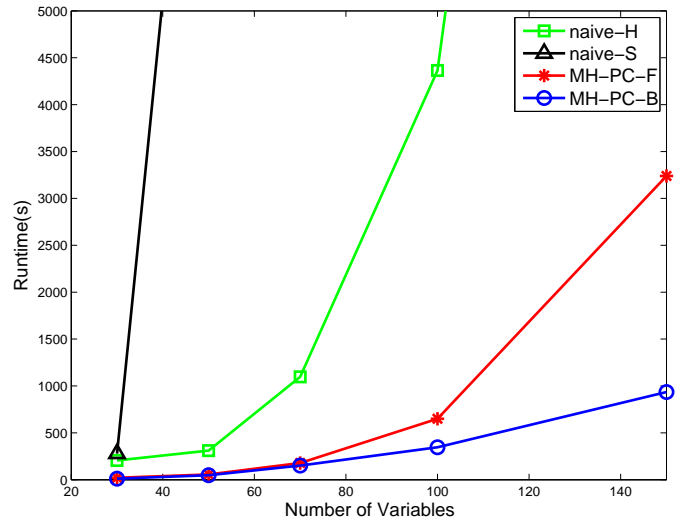}
            \caption{Scalability with number of variables - Census data}
            \label{fig:cen-var}
            \end{center}
        \end{figure}

        \begin{figure}[!b]
            \begin{center}
            \includegraphics[width=0.65\textwidth]{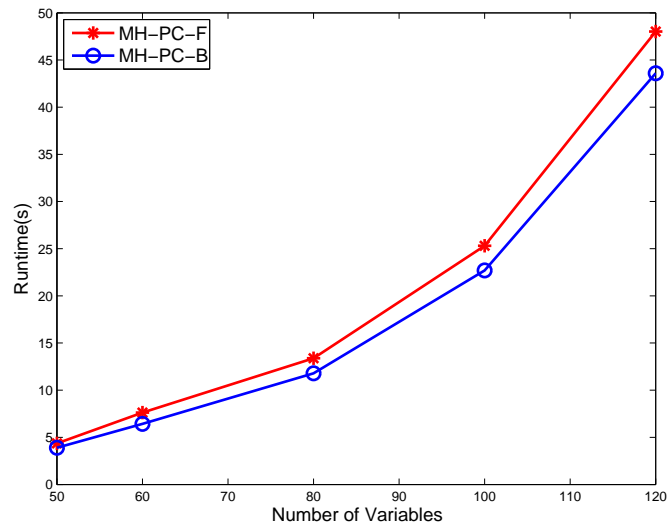}
            \caption{Scalability with number of variables - Synthetic data}
            \label{fig:syn-var}
            \end{center}
        \end{figure}

        We ran {Na\"{i}ve-H, Na\"{i}ve-S, CR-CS, CR-PA, MH-PC-F and MH-PC-B} with various data sets on the same computer with a 3.4 GHz quad-core CPU and 16 GB of memory.

        \begin{figure}[!t]
            \begin{center}
            \includegraphics[width=0.65\textwidth]{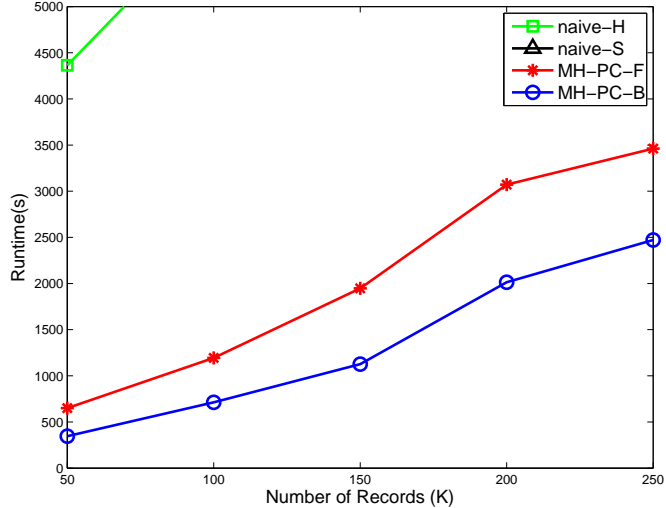}
            \caption{Scalability with umber of records - Census data}
            \label{fig:cen-record}
            \end{center}
        \end{figure}
								
        The running time of the algorithms on subsets of the Census data containing 30, 50, 70, 100 and 150 variables {with the same sample size (50K)} is shown in Figure \ref{fig:cen-var}. The two na\"{i}ve methods are much slower than MH-PC-F and MH-PC-B, and Na\"{i}ve-S is the most inefficient one. While the two na\"{i}ve methods do not scale {well with} the number of variables, the two proposed algorithms both {perform} good scalability.

        When applying the algorithms to the synthetic data sets containing different numbers of {variables}, both na\"{i}ve methods do not return results after 5 hours. So no results of na\"{i}ve methods are shown in Figure \ref{fig:syn-var}. From the figure, both proposed algorithms again scale well.

        We then ran the algorithms with 50K, 100K, 150K, 200K, and 250K samples respectively from the Census data set {with 100 variables selected randomly}, and the execution time of MH-PC-F and MH-PC-B is shown in Figure \ref{fig:cen-record}. {{No} results are obtained for Na\"{i}ve-S, and Na\"{i}ve-H also cannot handle data sets with more than 50K samples.} Similarly, MH-PC-B {is} more efficient and scalable than MH-PC-F.
		
        To summarise, MH-PC-F and MH-PC-B are much faster than the na\"{i}ve methods, and both proposed algorithms scale well in terms of the number of variables and number of samples. The experiments have also confirmed the discussions in Section \ref{subsectionMH-PC-F} that MH-PC-B can achieve higher efficiency than MH-PC-F.

\section{Conclusion} \label{sectionConclu}
    In practice, it is useful to identify a cause consisting of multiple variables, which individually are not causes of the target variable. However, finding such combined causes is challenging as the number of combined variables will increase exponentially with the increase of the number of individual variables. As far as we know, there has been very little work on discovering the combined causes, and the problem has not been studied in causal Bayesian network research either.

    In this paper, we have proposed two efficient algorithms to mine the combined causes from large data sets. The proposed algorithms are based on a well-designed local causal discovery method, the semi-interleaved HITON-PC {algorithm}, with the novel extensions for dealing with combined causes. Experiments have shown that the proposed algorithms can find single and combined causes with a low number of false discoveries from synthetic data sets, and discover many reasonable combined causes from real world data. Additionally, the algorithms have been shown to scale up well with respective to the number of variables and the number of samples with both synthetic and real world data.

    In the near future, we will apply the proposed algorithms to solving real world problems, such as investigating the mechanisms of gene regulation, for which there is evidence showing that many gene regulators work together to regulate their target genes.

\section{Acknowledgement}
    This work has been supported by Australian Research Council (ARC) Discovery Project Grant DP140103617.





\end{document}